\documentclass{article}

\usepackage{spconf,amsmath,graphicx,amssymb,hyperref}
\usepackage{microtype}

\usepackage{url}
\usepackage{booktabs}

\usepackage{lineno}

\usepackage{comment}
\usepackage{subcaption}
\usepackage{algorithm}
\usepackage{algpseudocode}
\usepackage{bm}
\usepackage{multirow}

\usepackage{colortbl}
\usepackage[normalem]{ulem}
\usepackage{algorithm}
\usepackage{algorithmicx}
\usepackage{algpseudocode}

\usepackage{enumitem}
\usepackage{makecell}
\useunder{\uline}{\ul}{}
\usepackage{booktabs}
\usepackage{multirow}
\usepackage{marvosym}
\usepackage[table]{xcolor}
\usepackage{amsmath} 

\title{Experience-Driven Dynamic Exits for LLMs with Reinforcement Learning}
\name{\begin{tabular}{c}
      Yanyu Zhu\textsuperscript{1,*}\thanks{* Equal contribution},
      Hoilam Pao\textsuperscript{1,*},
      Niu Hu\textsuperscript{3},
      Wei Guo\textsuperscript{3},
      Shaoxiong Zhan\textsuperscript{1},\\
      Boyu Lai \textsuperscript{4},
      Zitai Wang\textsuperscript{1},
      Yongqin Zeng\textsuperscript{1},
      Hai-Tao Zheng\textsuperscript{1,2,\Letter}\thanks{\Letter \ Corresponding to: Hai-Tao Zheng: zheng.haitao@sz.tsinghua.edu.cn.}\end{tabular}}

\address{\textsuperscript{1}Tsinghua Shenzhen International Graduate School \\
\textsuperscript{2}Pengcheng Laboratory, Shenzhen, China \\
\textsuperscript{3}Huawei Noah’s Ark Lab \\
\textsuperscript{4}Northwestern Polytechnical University \\
\{zhu-yy,bkl24\}@mails.tsinghua.edu.cn, zheng.haitao@sz.tsinghua.edu.cn}

\begin{document}
\ninept
\maketitle
\begin{abstract}
Large Language Models suffer from slow autoregressive inference. While self-speculative decoding accelerates this process, its efficiency is hampered by static configurations like fixed exit layers and speculation lengths. We reframe this optimization as a \textbf{Markov Decision Process} and propose \textbf{LEDE}, a framework that uses offline reinforcement learning. LEDE learns a policy to dynamically select the optimal exit layer and speculation length based on the local context of the generated sequence at each step, balancing computational cost and draft quality. Comprehensive evaluations on Llama-2 and Llama-3 models show LEDE achieves up to a $2.0\times$$\sim$$2.7\times$ speedup over autoregressive decoding and and provides an additional 17\% speedup over the static speculative baselines.

\end{abstract}



\begin{figure*}[htbp] 
\centering
\begin{minipage}[b]{1.0\linewidth}
  \centering
  \includegraphics[width=0.8\linewidth]{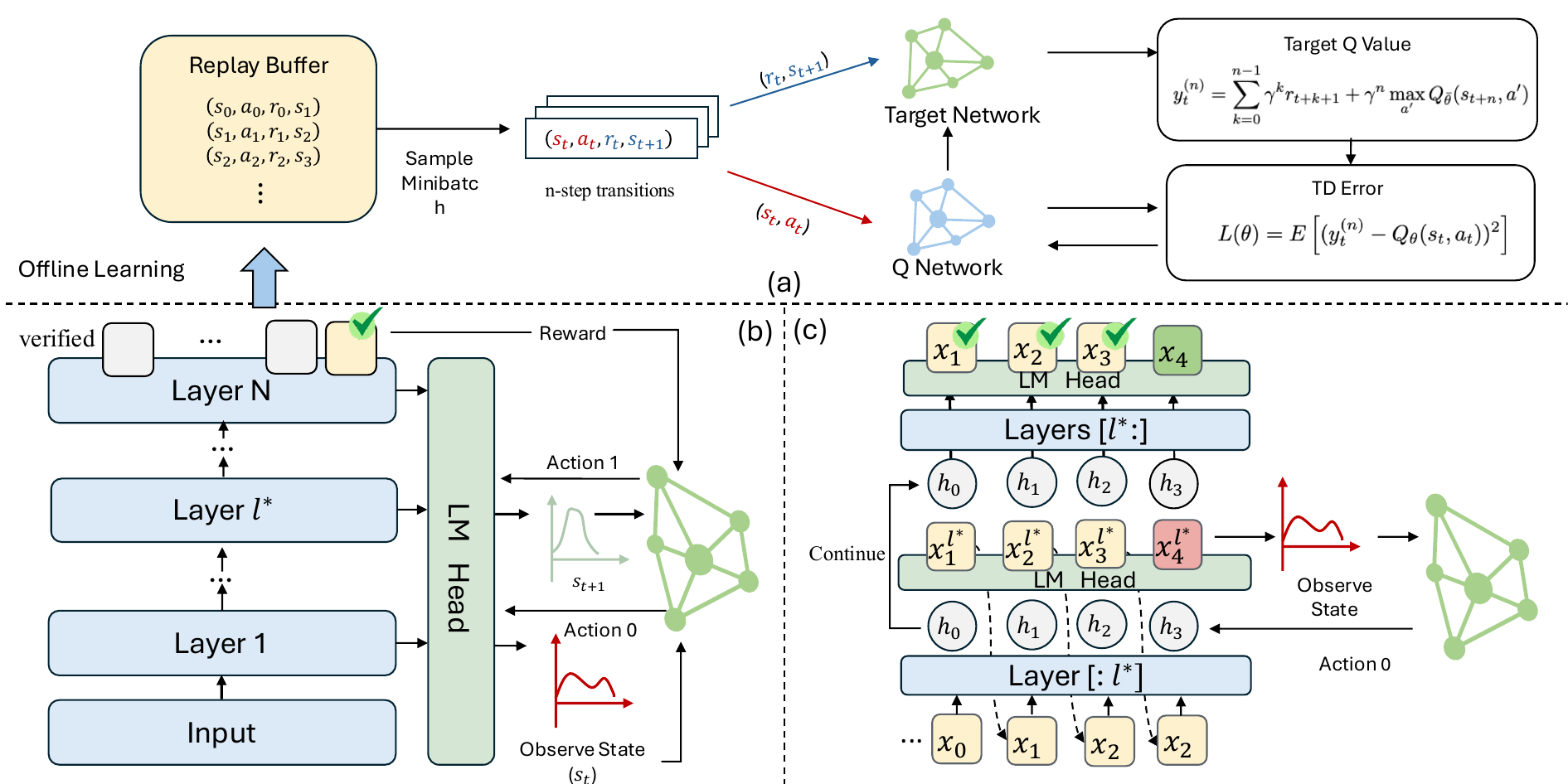}
\end{minipage}
\caption{The LEDE framework architecture. The framework consists of two stages: (a) an offline training phase where the Q-network learns an exit policy from a replay buffer, and an online inference phase (b-c). During inference, the trained agent first selects the optimal exit layer $l^*$ (b) and then adaptively determines the draft length at that layer (c).}
\label{fig:framework}
\end{figure*}

\section{Introduction}
\label{sec:intro}

Large Language Models (LLMs) \cite{llama2, gpt4, gemini} have become increasingly deep to handle various tasks including question answering, summarization, coding, and mathematical reasoning. While this depth enhances model capability, it also incurs considerable latency in the autoregressive decoding process, as every token must pass through the full stack of transformer layers. To mitigate this inefficiency, Self-Speculative Decoding (SSD) has been proposed, which exploits the model’s early layers as an internal “drafter” to generate candidate token sequences in parallel \cite{draft-and-verify}.

However, the efficiency of current SSD methods is fundamentally limited by their reliance on static configurations—a fixed draft depth and speculation length applied uniformly across the entire generation process \cite{layerskip}. This rigid approach is fundamentally misaligned with the existence of contextual sparsity in LLM inference \cite{not_all_layers, confidence_adaptive, dola, deja_vu}—not all tokens are equally difficult to predict. Consequently, existing methods, including those using simple heuristics \cite{draft_on_the_fly, lite, kangroo}, fail to optimally balance draft quality and speculation speed, leading to suboptimal performance.

To address this limitation, we introduce \textbf{LEDE} (Learning-based Dynamic Exit), a framework that formulates this dynamic control challenge as a \textbf{Markov Decision Process} (MDP) \cite{markov}. Specifically, LEDE applies offline reinforcement learning (RL) \cite{rl} to train a policy that dynamically controls both the draft depth and speculation length. At each step, the agent observes the model's internal state to assess generation difficulty and executes an action from its learned policy. This policy-driven approach replaces rigid, pre-defined rules with a sophisticated, adaptive strategy that generalizes from past inference experiences, enabling more robust and efficient decoding.

Our contributions are threefold: (1) We are the first to formulate the dynamic control of draft depth and speculation length in SSD as a MDP problem, applying RL to derive a control policy. (2) We design and implement LEDE, a framework that leverages offline RL and experience replay to learn its context-aware policy, overcoming the limitations of both static and simple heuristic-based configurations.(3) Our experiments demonstrate significant speedups of up to $\bm{2.7\times}$ over autoregressive generation and substantial efficiency gains over static baselines (up to \textbf{17\%}).

\label{sec:method}
\section{Related Work}
\label{sec:related work}
%
\subsection{Dynamic Computation Methods} To reduce the high cost of inference, dynamic computation methods adaptively adjust the compute allocated for each token. This is often achieved through early exiting, where tokens are predicted from an intermediate layer, or by skipping layers entirely. Seminal works like CALM and Mixture-of-Depths dynamically alter model depth based on token difficulty \cite{confidence_adaptive, mix_of_depth}. Other approaches explore unified layer skipping strategies \cite{unified_layer_skip}, contextual sparsity to activate only necessary sub-networks \cite{deja_vu, admtree, chen-etal-2025-dast}, or use k-NN search to optimize layer selection \cite{knn_sd}. These methods share the common goal of avoiding redundant computation for ``easy'' tokens. 

\subsection{Self-Speculative Methods}
Speculative decoding (SD) accelerates inference by using a faster and smaller draft model to generate tokens that are verified in parallel by the larger target model \cite{speculative_decoding, specinfer, eagle, unlocking}. A highly effective variant, SSD uses the target model's own early layers as the drafter, eliminating the need for external model \cite{draft-and-verify}. To better align the predictions of intermediate layers with those of the final layer for higher acceptance, some works \cite{kangroo, confidence_adaptive} train language head adapters to enable token prediction from intermediate depths. LayerSkip \cite{layerskip} employs a training recipe that combines layer droupout with a weighted early-exit loss to obtain robust early-exit capabilities. Our work builds upon this foundation but introduces a learning-based policy for dynamically controlling the drafting process.

\section{The LEDE Framework}

We introduce \textbf{Learning-based Dynamic Exit}, a framework that replaces static heuristics with a learned, adaptive policy to control the SSD process. As illustrated in Fig.~\ref{fig:framework}, our approach comprises two stages: an offline learning process, where an early-exit agent is trained from a repository of past inference experiences (Fig.~\ref{fig:framework}a), and an online inference process (Fig.~\ref{fig:framework}b-c), where the trained agent is deployed to make context-aware exit decisions. The following subsections detail each component of this framework.

\subsection{Markov Decision Process}

We formulate the dynamic selection of the exit layer $l^*$ as MDP framework. Unlike prior work that selects a fixed exit layer, our agent makes a series of decisions during the forward pass of a single generation step to determine the optimal depth for drafting.

\subsubsection{State Space}
The state observed by the agent is dynamic and depends on the computational depth. During the forward pass for a single generation step, upon reaching a candidate exit layer $l$, we extract a state vector, denoted as $s_l$. This vector is composed of several features derived from the model's internal token distribution at layer $l$, serving as compact, intrinsic signals that capture the model's confidence and uncertainty at that specific depth. These features are derived from the next-token probability distribution $P_t^{(l)}$ , which is computed from the hidden states at layer $l$. 

\textbf{Token Confidence ($C^l_k$).}
This is defined as the average negative log-probability of the top-k tokens from $P_t^{(l)}$:
$$
C^l_k = -\frac{1}{k}\sum^k_{j=1} logP_t^{(l)}(j)
$$
where $P_t^{(l)}(j)$ are the probabilities of the top-k predictions, with $k=5$ in our experiments. 

\textbf{Token Entropy ($H^l$).} It measures the uncertainty of the next-token prediction distribution at layer $l$ calculated as:
$$
H^l = -\sum_{j \in V} P_t^{(l)}(j)logP_t^{(l)}(j)
$$
where $V$ denotes the vocabulary set.

\textbf{Average Trace of Token Confidence ($C_{trace}$).}
It is defined as the average negative log-probability of the top-1 predicted token over the window:
$$
C_{trace} = -\frac{1}{T}\sum^T_{k=1} log P_{t-k}
$$
where $T$ denotes the window size and $P_{t-k}$ is the maximum probability in the output distribution at a previous step $t-k$, with $T=20$ in our experiments.






\subsubsection{Action Space}
At each candidate layer $l$, the policy Q-network takes the state $s_t$ as input and decides on an action $a_t \in \{continue, exit\}$. The \textit{exit}($a_t=1$) action designates the current layer $l$ as the chosen exit layer $l^*$ for the draft stage. The \textit{continue}($a_t=0$) action proceeds to the next candidate layer in the set. If the agent reaches the final candidate layer, it is forced to exit.
\subsubsection{Reward Function}
\label{sec:reward}
The reward function $R$ is designed to directly optimize for inference efficiency. For every exit at $l^*$, if the draft token is accepted, we give a positive reward for correctness and efficiency. Conversely, we give a negative reward for a rejected draft token. And we also give a small penalty for continuing to stimulate early exit. The reward function is defined as:
$$
R =
\begin{cases}
  1 + \frac{(L - l^*)}{L}, & \text{if } a_t = 1 \cap x^{l^*}_t = x^L_t, \\
  -2 + \frac{(L - l^*)}{L}, & \text{if } a_t = 1 \cap x^{l^*}_t \neq x^L_t, \\
  -0.01, & \text{if } a_t = 0.
\end{cases}
$$
where $L$ denotes the total number of layers in the model and $x^{l}_t$ represents the token predicted at layer $l$.

\subsection{Policy Optimization via Offline RL}
Our target is to train a Deep Q Network (DQN)\cite{q_learning} to optimize exit decisions. The key components of the framework are:

\textbf{Target Q network.} We employ a small MLP as our Q-network. It takes the state vector $s_t$ as input and outputs the Q-values, $Q(s_t, a_t)$ for the \textit{continue} and \textit{exit} actions. The action at the current layer is then selected greedily based on the maximum Q-value.

\textbf{Experience Collection and Replay.} As shown in Fig.~\ref{fig:framework}a, during the initial exploration phase, the policy agent continuously interacts with the environment and collects a diverse set of experiences. Each decision within the layer-wise forward pass is stored as an experience tuple $\left(s_t, a_t, r_t, s_{t+1}\right)$ in a replay buffer. Here, $s_t$ is the contextual state at a candidate layer $l$, $a_t$ is the action taken (\textit{continue} or \textit{exit}), and $s_{t+1}$ is the state at the subsequent candidate layer. The reward $r_t$ is assigned according to the reward function defined in ~\ref{sec:reward}. This replay buffer of experiences provides the comprehensive dataset for offline policy optimization.

\textbf{Offline Policy Update.} The Q-network is trained offline by sampling mini-batches of experiences from the replay buffer. The policy is optimized by minimizing the n-step Temporal Difference (TD) error \cite{n-step-returns}. This is achieved by minimizing the mean square error between the predicted Q-value and a target value $y_t$. The loss function is defined as:
$$
L(\theta) = \mathbb{E}\left[(y_t^{(n)} - Q_\theta(s_t, a_t))^2 \right]
$$
where the n-step target return $y_t^{(n)}$ is the sum of discounted rewards for n steps plus the discounted value of the state reached after n steps, estimated by a stable target network $Q_{\bar{\theta}}$:
$$
y_t^{(n)} = \sum^{n-1}_{k=0} \gamma^k r_{t+k+1} + \gamma^n \max_{a'}Q_{\bar{\theta}}(s_{t+n}, a')
$$
where $\gamma$ is the discount factor and $a'$ is the greedy action from the target network $Q_{\bar{\theta}}$, with $\gamma=0.99$ and $n=3$ in our experiments.

\subsection{Dynamic Exit Layer and Adaptive Drafting}
\label{sec:selection_and_drafting}

Our trained policy agent dynamically controls both the exit layer and draft length, as depicted in the online inference stage of our framework. First, the agent selects the optimal exit layer $l^*$ by making an ``exit'' decision during a layer-wise evaluation in the initial forward pass (Fig.~\ref{fig:framework}b). Subsequently, it generates draft tokens starting from $l^*$. After each draft, the agent re-evaluates the state: an ``exit'' action signals high confidence and continues the drafting process, while a ``continue'' action indicates low confidence and immediately terminates drafting for verification (Fig.~\ref{fig:framework}c). This dual-control mechanism co-optimizes computational depth and speculation length to maximize inference speed.

\begin{table*}[h!]
\centering
\caption{Comparison of LEDE against baselines across different models and a wide range of tasks using LLaMA2 and LLaMA3 models. Speedup is measured relative to autoregressive decoding. Best results are in \textbf{bold}, and second-best results are \underline{underlined}.}
\label{tab:main_results}
\resizebox{\textwidth}{!}{%
\begin{tabular}{l l ccccc ccccc ccccc ccccc}
\toprule
\multirow{2}{*}{\textbf{Model}} & \multirow{2}{*}{\textbf{Method}} & \multicolumn{5}{c}{\textbf{CNN/DM (Summarization)}} & \multicolumn{5}{c}{\textbf{CNN/DM (Lang)}}&  \multicolumn{5}{c}{\textbf{TOPv2 (Instruction)}} & \multicolumn{5}{c}{\textbf{Alpaca (Instruction)}}\\
\cmidrule(lr){3-7} \cmidrule(lr){8-12} \cmidrule(lr){13-17} \cmidrule(lr){18-22} 
& & E & d  & Acc. Rate & Speedup & R-L & E & d & Acc. Rate & Speedup & R-L & E & d  & Acc. Rate & Speedup & R-L & E & d  & Acc. Rate & Speedup & R-L  \\

\midrule
\multirow{5}{*}{LLaMA-3.2-1B} & AR & - & - & - & $1.00\times$ & 0.117 & - & - &  - & $1.00\times$  & 0.108 & - & - &  - & $1.00\times$ & 0.068  & - & - &  - & $1.00\times$ & 0.153 \\
& LS & 4.00 & 4.00 & 0.632  & \underline{$1.60\times$}  & 0.113 & 4.00 & 4.00 & 0.740 & \underline{$1.81\times$} & 0.108 & 4.00 & 4.00 & 0.771 & \underline{$1.86\times$}  & 0.068 & 4.00 & 4.00 & 0.751 & \underline{$1.73\times$} & 0.153 \\
& LITE & 10.36 & 11.83 & 0.922 & $1.07\times$ & 0.117 & 9.77 & 11.63 & 0.844 & $1.30\times$  & 0.108 & 9.27 & 11.76 & 0.863 & $1.45\times$ & 0.680  & 10.07 & 11.59 & 0.849 & $1.09\times$ & 0.155  \\
& DV & 6.00 & 5.61 & 0.882 & $1.44\times$  & 0.116 & 6.00 & 7.26 & 0.868 & $1.65\times$  & 0.108 & 3.00 & 6.94  & 0.905 & $1.78\times$ & 0.068 & 6.00 & 6.05  & 0.867 & $1.23\times$  & 0.156 \\
\rowcolor{cyan!5}
\cellcolor{white}& LEDE & \textbf{5.70} & \textbf{6.84}  & \textbf{0.924} & $\bm{2.04\times}$ & 0.116 & \textbf{4.62} & \textbf{6.40}  & \textbf{0.881} & $\bm{2.28\times}$  & 0.107 & \textbf{6.88} & \textbf{6.40}  & \textbf{0.911} & $\bm{1.98\times}$  & 0.068 & \textbf{3.96} & \textbf{4.70}  & \textbf{0.867} & $\bm{2.04\times}$  & 0.156  \\
\midrule
\multirow{5}{*}{LLaMA-2-7B} & AR & - & - & - & $1.00\times$  & 0.194 & - & -  & - & $1.00\times$ & 0.225 & - & - & - & $1.00\times$ & 0.095 & - & - & - & $1.00\times$  & 0.207 \\
& LS & 4.00 & 4.00 &  0.254 & $1.22\times$  & 0.193 & 4.00 & 4.00 & 0.553 & $1.95\times$  & 0.225 & 4.00 & 4.00  & 0.732 & \underline{$2.47\times$} & 0.094 & 4.00 & 4.00  & 0.724 & \underline{$2.37\times$} & 0.206 \\

& LITE & 15.39 & 11.82  & 0.955 & $1.54\times$ & 0.193 & 19.17 & 11.88  & 0.806  & $1.32\times$ & 0.225 & 18.00 & 0.881  & 10.417 & $1.52\times$ & 0.094 & 18.49 & 11.83  & 0.881 & $1.37\times$ & 0.204 \\

& DV & 8.00 & 7.57  & 0.858 & \underline{$2.09\times$}  & 0.194 & 8.00 & 5.60 & 0.865 & \underline{$2.13\times$}  & 0.225 & 8.00 & 6.29 & 0.779 & $1.94\times$ & 0.094 & 8.00 & 7.57  & 0.893 & $2.35\times$  & 0.205\\
\rowcolor{cyan!5}
\cellcolor{white}& LEDE & \textbf{6.82} & \textbf{7.82}  & \textbf{0.923} & $\bm{2.64\times}$  & 0.193 & \textbf{7.05} & \textbf{6.20} & \textbf{0.904} & $\bm{2.72\times}$   & 0.225 & \textbf{8.72} & \textbf{6.72} & \textbf{0.900} & $\bm{2.23\times}$  &   0.094 & \textbf{7.40} & \textbf{9.40}  & \textbf{0.981} & $\bm{2.58\times}$  & 0.205 \\
\midrule

\multirow{5}{*}{LLaMA-2-13B} & AR & - & - & - & $1.00\times$ & 0.2  & - & - &  - & $1.00\times$ & 0.248 & - & - & - & $1.00\times$  & 0.157  & - & - & - & $1.00\times$  & 0.246\\
& LS & 8.00 & 4.00  & 0.743 & \underline{$2.27\times$}  & 0.199 & 8.00 & 4.00  & 0.711 & \underline{$2.12\times$}  & 0.247 & 8.00 & 4.00 & 0.808 & $2.28\times$  & 0.156 & 8.00 & 4.00  & 0.799 & \underline{$2.25\times$} & 0.246 \\
& LITE & 25.42 & 11.73 & 0.817 & $1.39\times$  & 0.199 & 26.58 & 11.86 & 0.780 & $1.21\times$  & 0.246 & 24.53 & 11.85 & 0.868 & $1.46\times$ & 0.156 & 25.00 & 11.85 & 0.861 & $1.27\times$  & 0.246 \\

& DV & 15.00 & 6.64 & 0.958 & $2.23\times$  & 0.2 & 15.00 & 6.25 &  0.925 & $2.11\times$  & 0.246 & 15.00 & 5.98 & 0.915 & \underline{$2.43\times$}  & 0.153 & 15.00 & 5.64 &  0.944 & $2.13\times$  & 0.246 \\
\rowcolor{cyan!5}
\cellcolor{white} & LEDE & \textbf{9.36} & \textbf{5.19} & \textbf{0.928} & $\bm{2.55\times}$  & 0.199 & \textbf{9.69} & \textbf{4.63} & \textbf{0.969} & $\bm{2.57\times}$  & 0.246 & \textbf{12.73} & \textbf{8.06}  & \textbf{0.955} & $\bm{2.56\times}$ & 0.155 & \textbf{10.70} & \textbf{7.17}  & \textbf{0.969} & $\bm{2.51\times}$ & 0.246 \\
\bottomrule
\end{tabular}%
}
\end{table*}

\begin{table}[h]
    \centering
    \caption{Speedup comparison on the code generation task. LEDE outperforms all baselines, achieving 2.18$\times$ and 2.07$\times$ speedups on CodeLLaMA-7B and CodeLLaMA-34B, respectively.}

    \scalebox{0.7}{
        \begin{tabular}{llcccc}
        \toprule
        \multirow{2}{*}{\textbf{Model}}&\multirow{2}{*}{\textbf{Method}} & \multicolumn{4}{c}{\textbf{HumanEval}} \\
         \cmidrule(lr){3-6} 
        & & E & d & Acc. Rate & Speedup \\
        \midrule
        \multirow{5}{*}{CodeLLaMA-7B}&AR & - & - & - & $1.00\times$  \\
        &LS & 7.00 & 6.00 & 0.645 & $1.94\times$ \\
        &LITE & 20.24 & 11.72 & 0.838 & $1.30\times$   \\
        &DV & 8.00 & 5.90 & 0.790 & $2.02\times$  \\
        &LEDE & \textbf{8.35} & \textbf{5.93} & \textbf{0.748} & $\bm{2.18\times}$  \\
        \midrule
        \multirow{5}{*}{CodeLLaMA-34B}&AR & - & - & - & $1.00\times$  \\
        &LS & 6.00 & 12.00 & 0.272 & $1.61\times$ \\
        &LITE& 30.79 & 11.47 & 0.826 & $1.25\times$\\
        &DV& 12.00 & 5.57 & 0.765 & $1.88\times$\\
        &LEDE& \textbf{14.57} & \textbf{4.41} & \textbf{0.864} & $\bm{2.07\times}$\\
        \bottomrule
        \end{tabular}
    } 
    \label{tab:codellama}
\end{table}

\section{Experiment}
\subsection{Implementaion Details}

\textbf{Training Hyperparameters.} The model is trained using the Adam optimizer \cite{adam} with a learning rate of $\alpha = 6.25 \times 10^{-5}$. We balance exploration and exploitation by incorporating parameter-space noise via factorized Gaussian NoisyLinear Layers \cite{noisy_net} initialized with a value of $\sigma_0=0.1$ in our Q-network. To populate the experience replay buffer, an initial exploration phase is conducted for 200 episodes before any learning begins. Subsequently, a learning update is performed every 4 agent steps using a mini-batch of 32 transitions, and the target network is updated via a copy of the online network's parameters every 512 steps. 


\textbf{Evaluation Setup.} We evaluate our framework on a range of models to demonstrate scalability: LLaMA3.2-1B, LLaMA2-7B, LLaMA2-13B, CodeLLaMA-7B and CodeLLaMA-34B. All these models are continually pretrained by LayerSkip\cite{layerskip}. The evaluation datasets span five distinct tasks: instruction following on Alpaca\cite{taori2023alpaca} and TOPv2\cite{chen2020low}, language modeling and summarization on CNN/DailyMail(CNN/DM)\cite{nallapati2016abstractive}, code generation on HumanEval\cite{chen2021evaluating}. We employ sampling decoding with a temperature of 0.6 for the language modeling task and greedy decoding for all other tasks. All evaluations use speculative sampling\cite{speculative-sampling} as the acceptance strategy, with a maximum generation length of 512 and a batch size of 1. Our implementation is built upon the LayerSkip\cite{layerskip} open-source codebase. All experiments were conducted on an NVIDIA A100 (SXM4-80GB) with CUDA 12.2 and PyTorch 2.6.0.



\textbf{Baselines.} We compare LEDE against four representative baselines. \textbf{Autoregressive (AR)} stands for the standard, non-accelerated decoding method as the fundamental baseline (1.00$\times$ speedup). \textbf{LayerSkip (LS)} \cite{layerskip} represents the state-of-the-art of SSD method with fixed draft depth $E$ and a static speculation length $D$. \textbf{LITE}\cite{lite} employs a rule-based dynamic early-exit method that uses predefined, layer-specific confidence thresholds to trigger an exit. \textbf{Draft \& Verify (DV)}\cite{draft-and-verify} uses a static exit layer but controls the draft length with an adaptive confidence threshold to exit drafting early. For the dynamic speculative strategies, we restrict the maximum speculation length to 12 per speculation round.

\textbf{Evaluation Metrics.} We report the four key metrics for LEDE evaluations: the \textbf{Average Speculation Length (d)}, the \textbf{Acceptance Rate (Acc. Rate)}, which measure speculation efficiency, the \textbf{Average Exit Layer (E)}, which indicates the computational depth used; and wall-clock \textbf{Speedup} relative to autoregressive decoding. Additionally, \textbf{R-L (Rouge-L)} is used to verify generation quality.


\subsection{Main Results}

The performance comparison across LLaMA model sizes and text generation tasks is presented in Table~\ref{tab:main_results} LEDE robustly outperforms all baselines by learning a superior trade-off between computational depth and speculation efficiency. Our method achieves an average speedup of $\bm{2.32\times}$ and up to $\bm{2.72\times}$, significantly surpassing static and heuristic-based methods. Table~\ref{tab:codellama} reports the acceleration of code generation tasks with CodeLLaMA-7B and 34B. The core of the acceleration lies in its ability to dynamically co-optimize draft deepth (E), draft length (d) and token acceptance rate (Acc. Rate). In contrast, the limitations of alternative approaches are clear. The performance of LayerSkip(LS) is highly sensitive to its static (E, d) configuration, which fails to generalize effectively across tasks. The rule-based adaptive strategies are also constrained by their predefined thresholds or hyperparameters, leading to sub-optimal trade-offs and lower overall speedup. By replacing rigid heuristics with a finer-grained control policy, LEDE provides a more robust and effective solution for accelerating LLM inference across diverse models and tasks.

\subsection{Ablation Study}
To isolate the contribution of each component of our dual-control policy, we conducted an ablation study on LLaMA2-7B. We tested two variants: one without dynamic exit layer selection \textbf{(LEDE w/o Dynamic Exit)} and another without adaptive drafting length \textbf{(LEDE w/o Adaptive Drafting)}. As shown in Table~\ref{tab:ablation}, removing either component significantly degrades performance, resulting in lower acceptance rates and reduced speedups compared to the full \textbf{LEDE} model. This confirms that both mechanisms are integral to our framework's success and have a synergistic effect on overall efficiency.






\begin{table}[h]
\centering
\caption{Ablation study of LEDE's dynamic components on LLaMA2-7B.}
\label{tab:ablation}
\begin{tabular}{lcccc}
\toprule
\textbf{Configuration} & E & d & Acc. Rate & Speedup \\
\midrule
\textbf{LEDE (Full)} & \textbf{7.39} & \textbf{7.84} & \textbf{0.858} & $\bm{2.7\times}$ \\
\midrule
w/o Adaptive Drafting& 7.2 & 6.0  & 0.700 & 2.04$\times$ \\
w/o Dynamic Exit & 6.0  & 6.87 & 0.690 & 1.99$\times$ \\
\bottomrule
\end{tabular}
\end{table}

\begin{figure}[htb]
\centering


\begin{minipage}[b]{0.8\linewidth}
  \centering
  \includegraphics[width=\linewidth]{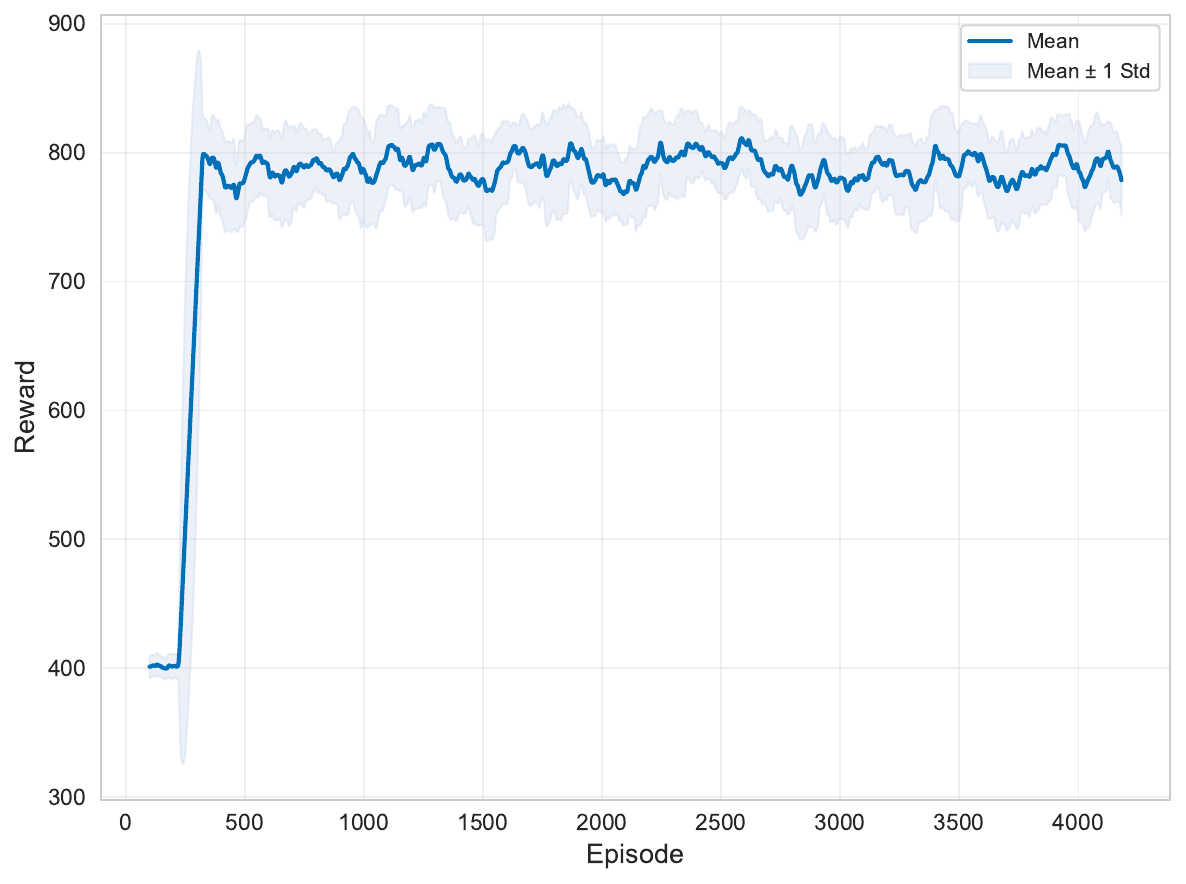}
  \centerline{(a) Reward}
\end{minipage}

\begin{minipage}[b]{0.48\linewidth}
  \centering
  \includegraphics[width=\linewidth]{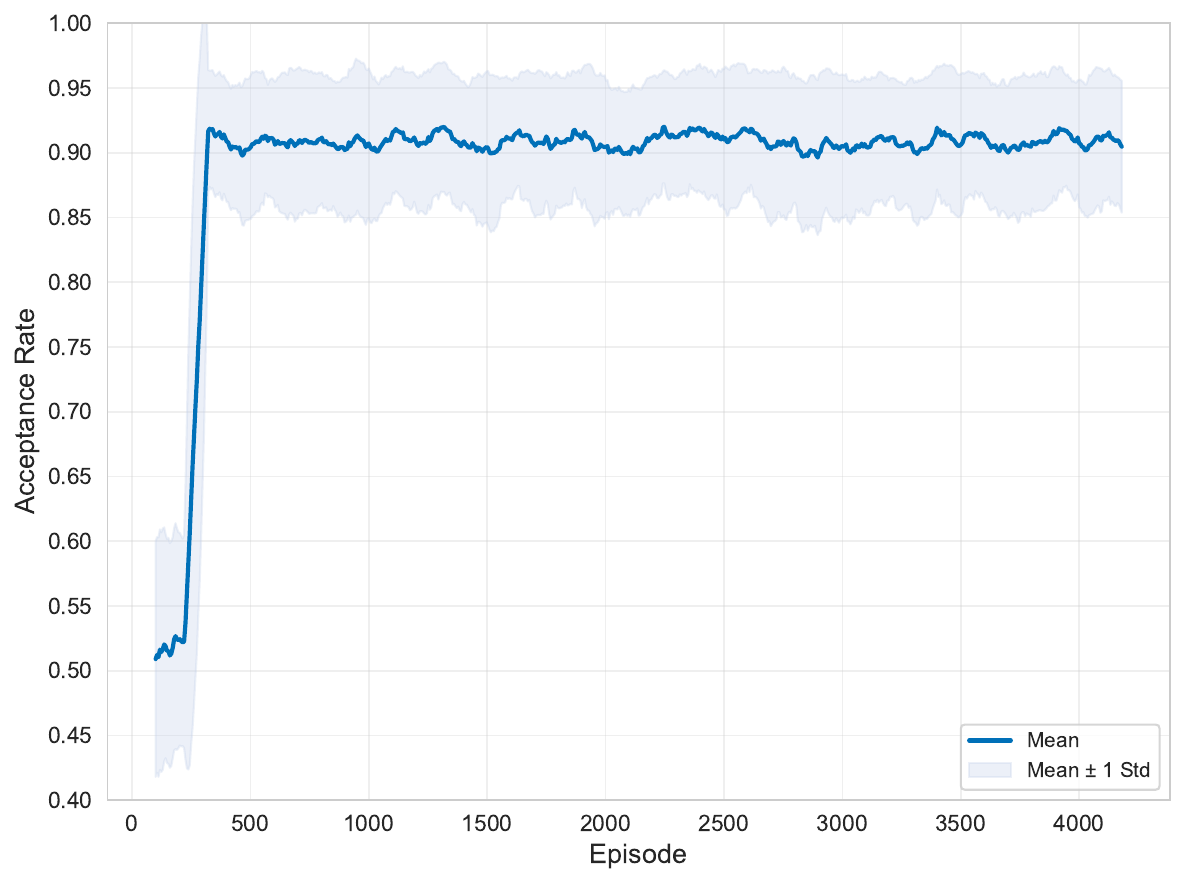}
  \centerline{(b) Average Acceptance Rate}
\end{minipage}
 \hfill
\begin{minipage}[b]{0.48\linewidth}
  \centering
  \includegraphics[width=\linewidth]{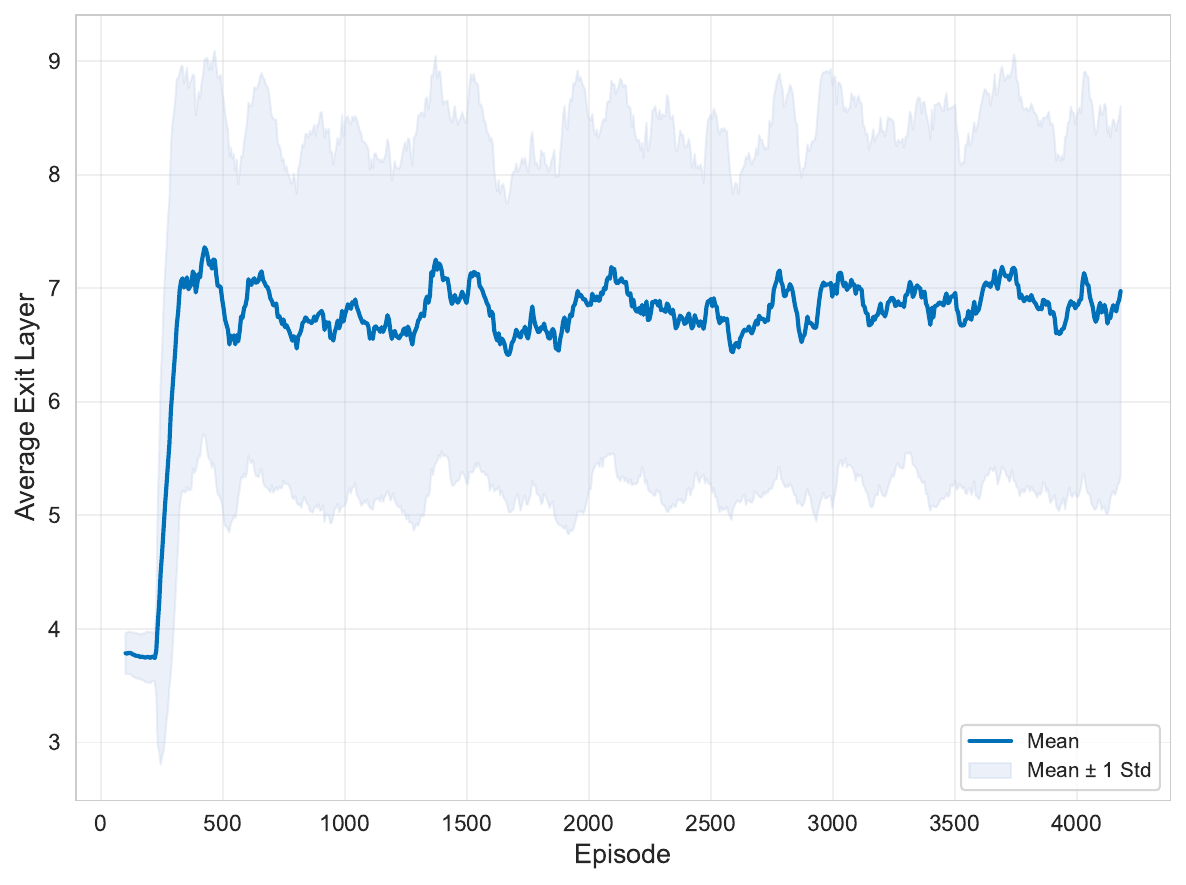}
  \centerline{(c) Average Exit Layer}
\end{minipage}

\caption{Policy learning dynamics.}
\label{fig:reward}
\end{figure}

\subsection{Learning Dynamic Analysis}
Fig.~\ref{fig:reward} illustrates the learning dynamic across the training episodes. After an initial exploration phase (the first 200 episodes), the agent rapidly converges to a stable policy, as shown by the reward plateau in Fig.~\ref{fig:reward}a. This convergence is driven by the agent learning to balance computational depth with speculation quality: the average acceptance rate (Fig.~\ref{fig:reward}b) climbs to a stable $\sim0.9$, while the average exit layer (Fig.~\ref{fig:reward}c) settles at an optimal depth of $\sim7$. These dynamics confirm that our learning framework effectively discovers a robust policy that optimizes the efficiency-quality tradeoff.

\section{Conclusion}
We introduced LEDE, a novel framework that employs offline reinforcement learning to dynamically control SSD. By learning a policy that dynamically co-optimizes draft depth and speculation length, LEDE achieves significant $2.0\sim2.7\times$ speedups over autoregressive decoding across diverse models and tasks. Our approach represents a conceptual shift from rigid heuristics toward adaptive, policy-driven inference. Future work will focus on scaling these benefits to models at the 70B scale and beyond. We hope this work highlights the potential of reinforcement learning to build more intelligent and computationally efficient LLM systems.

\section{Acknowledgment}
This research is supported by National Natural Science Foundation of China (Grant No.62276154);the Natural Science Foundation of Guangdong Province (Grant No.2024TQ08X729);Basic Research Fund of Shenzhen City (Grant No.JCYJ20240813112009013 and GJHZ20240218113603006);The Major Key Project of PCL for Experiments and Applications (Grant No.PCL2024A08).
\bibliographystyle{IEEEbib}
\bibliography{strings,refs}



\end{document}